\documentclass[10pt]{article}

\usepackage[margin=0.75in]{geometry}
\usepackage[T1]{fontenc}
\usepackage[utf8]{inputenc}
\usepackage{lmodern}
\usepackage{microtype}
\usepackage{graphicx}
\usepackage{booktabs}
\usepackage{amsmath,amssymb}
\usepackage{multirow}
\usepackage{hyperref}
\usepackage[numbers,sort&compress]{natbib}
\usepackage{xcolor}
\usepackage{caption}

\hypersetup{
    colorlinks=true,
    linkcolor=blue,
    citecolor=blue,
    urlcolor=blue,
    pdftitle={Matched-Learning-Rate Analysis of Attention Drift and Transfer Retention in Fine-Tuned CLIP},
    pdfauthor={Ruize Xia},
}

\title{Matched-Learning-Rate Analysis of Attention Drift and Transfer Retention in Fine-Tuned CLIP}
\author{Ruize Xia\\Independent Researcher\\\texttt{xiaruize0911@gmail.com}}
\date{April 2026}

\begin{document}
\maketitle

\begin{center}
\small Preprint prepared for arXiv. Companion code and analysis assets are available at \url{https://github.com/xiaruize0911/Attention_Collapse_in_CLIP_Fine-tuning_repo}.
\end{center}
\vspace{-1em}

\begin{abstract}
CLIP adaptation can improve in-domain accuracy while degrading out-of-domain transfer, but comparisons between Full Fine-Tuning (Full FT) and LoRA are often confounded by different learning-rate conventions. We study how adaptation method and optimization scale jointly shape attention drift and transfer retention in CLIP using a controlled matched-learning-rate comparison of Full FT and LoRA. The completed matrix contains 80 runs on CLIP ViT-B/32 across EuroSAT and Oxford-IIIT Pets, spanning four shared learning rates ($10^{-6}$, $5{\times}10^{-6}$, $10^{-5}$, $5{\times}10^{-5}$) and five seeds, and evaluates attention-drift metrics, best validation accuracy, and adapter-aware CIFAR-100 zero-shot accuracy. Learning rate strongly modulates structural change: on EuroSAT, Full FT moves from mild entropy broadening at $10^{-6}$ to marked contraction at $5{\times}10^{-5}$, whereas LoRA remains entropy-positive across the full matched grid. At matched learning rates, LoRA preserves substantially more zero-shot transfer than Full FT, averaging $45.13\%$ versus $11.28\%$ CIFAR-100 accuracy on EuroSAT and $58.01\%$ versus $8.54\%$ on Pets. Oxford-IIIT Pets also reveals a regime effect: low-learning-rate LoRA underfits in-domain, so method-only averages can obscure when LoRA becomes competitive. Supporting rollout, patch-to-patch, and CKA analyses are directionally consistent with the controlled matrix. Overall, matched-learning-rate evaluation materially changes the interpretation of Full FT versus LoRA, and attention drift is most useful as a descriptive diagnostic of representation preservation rather than a causal explanation of transfer behavior.
\end{abstract}

\section{Introduction}
Contrastive vision--language pretraining makes CLIP a strong starting point for downstream adaptation, but adaptation can trade in-domain gains against the transfer behavior that makes CLIP attractive in the first place \citep{radford2021learning}. For practitioners, the question is therefore not only whether fine-tuning improves task accuracy, but also how much of the pretrained model's out-of-domain capability survives the adaptation process. Prior work shows that full fine-tuning can distort pretrained representations and degrade robustness under shift, while parameter-efficient methods such as LoRA often preserve more of the original model \citep{kumar2022finetuning,wortsman2022robust,hu2022lora}.

Attention maps provide one descriptive lens on this trade-off in Vision Transformers, where spatial information routing is mediated by self-attention across tokens \citep{dosovitskiy2021image}. We do not treat attention as a causal explanation of model predictions \citep{jain2019attention,wiegreffe2019attention}. Instead, we use \emph{attention drift}---changes in CLS-to-patch attention support relative to the pretrained encoder---as a compact measurement of structural change during adaptation.

A recurring difficulty in comparisons between Full FT and LoRA is that the two methods are often trained under different learning-rate conventions. When optimization scales differ, apparent method effects mix with update-magnitude effects. The central question of this paper is therefore straightforward: under matched optimization scales, how do Full FT and LoRA differ in attention drift, in-domain performance, and out-of-domain transfer retention?

We answer this question with a completed 80-run matched-learning-rate study of CLIP ViT-B/32 across EuroSAT and Oxford-IIIT Pets. Full FT and LoRA share the same four-learning-rate grid and five random seeds, and we evaluate attention-drift metrics, best validation accuracy, and adapter-aware CIFAR-100 zero-shot accuracy. Rollout, patch-to-patch, CKA, regularization, backbone, and rank-ablation analyses are included as supporting evidence that broadens the empirical picture without displacing the main controlled matrix.

The contributions of this paper are fourfold:
\begin{itemize}
    \item We present a completed 80-run matched-learning-rate comparison of Full FT and LoRA on CLIP ViT-B/32 across EuroSAT and Oxford-IIIT Pets.
    \item We show that learning rate strongly modulates attention drift, and that method comparisons change materially when optimization scale is controlled.
    \item We find that LoRA generally preserves substantially more zero-shot transfer than Full FT at matched learning rates.
    \item We show that broader retained attention support is positively associated with transfer retention, while treating attention as a descriptive diagnostic rather than a stand-alone causal explanation.
\end{itemize}

\section{Related Work}
\subsection{CLIP Adaptation}
CLIP established large-scale image--text pretraining as a strong basis for transfer learning across diverse downstream tasks \citep{radford2021learning}. A substantial subsequent literature then examined what happens when zero-shot models are adapted rather than used directly. In particular, \citet{kumar2022finetuning} showed that full fine-tuning can distort pretrained features and underperform under distribution shift, while \citet{wortsman2022robust} argued that robust adaptation of zero-shot models requires more care than simply optimizing for in-domain accuracy. These studies motivate the central question of the present paper: if downstream adaptation changes transfer behavior, what measurable internal structural changes accompany that shift?

Our work differs from this prior line in emphasis. The studies above focus primarily on predictive robustness and representation quality at the output level. We instead examine a structural correlate inside the visual encoder, namely the drift of attention heatmaps relative to the pretrained baseline. The paper therefore complements, rather than replaces, robustness-oriented analyses of CLIP adaptation.

\subsection{Attention in Vision Transformers}
The visual backbone of CLIP follows the Vision Transformer formulation of \citet{dosovitskiy2021image}, in which token interactions are mediated by multi-head self-attention. That architecture makes it natural to study how adaptation changes information routing across depth. Prior work on attention flow, especially attention rollout \citep{abnar2020quantifying}, provides one concrete mechanism for summarizing layerwise attention patterns beyond any single matrix. This broader literature supports the idea that attention can be analyzed as a structural property of the model even when it is not interpreted as a complete explanation of prediction behavior.

At the same time, the interpretability status of attention remains contested. \citet{jain2019attention} argued that attention weights should not automatically be equated with explanations, whereas \citet{wiegreffe2019attention} showed that attention can still support informative analysis when its role is framed more carefully. We adopt that more conservative position. In this paper, attention heatmaps are treated as descriptive measurements of how downstream optimization reshapes the visual encoder, not as stand-alone causal explanations of model decisions.

\subsection{Parameter-Efficient Adaptation}
LoRA introduced a low-rank parameterization of downstream updates that greatly reduces the number of trainable parameters while preserving much of the base model \citep{hu2022lora}. Although LoRA was originally developed and evaluated in language modeling settings, the broader idea of parameter-efficient adaptation has clear relevance for vision--language models: if the update space is constrained, one may expect the adapted model to remain closer to the pretrained representation geometry. This intuition is especially relevant for CLIP, where preserving pretrained transfer behavior is often as important as maximizing in-domain accuracy.

The present paper builds on that motivation but introduces a stronger experimental control than most method-level comparisons. Rather than assuming that any observed Full-FT-versus-LoRA gap is attributable to the adaptation method alone, we explicitly hold the learning-rate grid fixed. This control is important because optimization scale can itself induce substantial representational and structural drift.

\subsection{Study Positioning}
Taken together, the prior literature leaves an important gap. CLIP adaptation work shows that robustness can degrade after fine-tuning, transformer-analysis work provides tools for studying internal attention structure, and LoRA offers a mechanism for constraining adaptation. What is less well established is how these pieces fit together under a controlled comparison that disentangles adaptation method from learning rate.

Our contribution is a controlled matched-learning-rate evaluation of attention drift in which method, learning rate, transfer retention, and layerwise structure are analyzed jointly. The supplementary analyses broaden the empirical picture, but the main claims remain anchored in the completed controlled matrix.

\section{Experimental Setup}
\subsection{Controlled Matched-Learning-Rate Design}
All experiments use CLIP ViT-B/32 (\texttt{openai/clip-vit-base-patch32}) with eager attention enabled so that per-head attention maps can be extracted directly from the vision encoder \citep{radford2021learning,dosovitskiy2021image}. The controlled matrix spans two downstream datasets, EuroSAT \citep{helber2019eurosat} and Oxford-IIIT Pets \citep{parkhi2012cats}, and evaluates transfer retention on CIFAR-100 \citep{krizhevsky2009learning}. We compare two adaptation methods:
\begin{itemize}
    \item Full FT, which updates the full visual encoder and classifier head; and
    \item LoRA r=8, which adapts the visual attention stack through low-rank updates while retaining a much smaller trainable parameter set \citep{hu2022lora}.
\end{itemize}

The matrix is fully crossed over four shared learning rates ($10^{-6}$, $5{\times}10^{-6}$, $10^{-5}$, $5{\times}10^{-5}$) and five seeds (7, 11, 19, 42, 123), yielding 80 completed runs: $2$ datasets $\times$ $2$ methods $\times$ $4$ learning rates $\times$ $5$ seeds.

The code, training scripts, and analysis pipeline for the paper are available in a \href{https://github.com/xiaruize0911/Attention_Collapse_in_CLIP_Fine-tuning_repo}{public repository}. The companion release includes the main matched-learning-rate experiments together with the supporting analyses discussed below.

\section{Metrics}
For each run we compare the adapted model against the pretrained baseline on a fixed, class-balanced evaluation protocol. Structural metrics are computed on in-domain validation images, while out-of-domain data are reserved for transfer evaluation. We focus on CLS-to-patch attention as a compact summary of spatial support in the visual encoder. The primary structural measurements are:
\begin{itemize}
    \item \textbf{CLS-to-patch entropy}, where higher values indicate broader attention support;
    \item \textbf{ERF@0.95}, the fraction of tokens needed to accumulate 95\% of attention mass \citep{luo2017erf};
    \item \textbf{Gini concentration}, where higher values indicate more concentrated attention; and
    \item \textbf{Head diversity}, based on pairwise cosine dissimilarity across heads.
\end{itemize}

We report each metric as the percent change from the pretrained baseline. We complement the structural measurements with best validation accuracy on the downstream task and adapter-aware CIFAR-100 zero-shot accuracy. The transfer metric is evaluated with the adapted image branch active so that LoRA checkpoints are not scored through a base-model-only path.

\section{Main Controlled Results}
\subsection{Learning-Rate Effects}
Figure~\ref{fig:lr-panels} summarizes the controlled matrix. Several consistent patterns emerge.

On EuroSAT, Full FT is highly sensitive to optimization scale. At the smallest learning rate, it mildly broadens attention ($+1.83\%$ entropy, $+2.42\%$ ERF on average across seeds), but as the learning rate increases it becomes progressively more contractive, reaching $-3.99\%$ entropy and $-8.25\%$ ERF at $5{\times}10^{-5}$. LoRA shows a distinct pattern: it remains entropy-positive across the full matched grid, with mean entropy shifts between $+0.68\%$ and $+1.50\%$. The in-domain accuracy gap is largest at the smallest learning rate and narrows substantially at larger values. By $5{\times}10^{-5}$, LoRA reaches $98.83\%$ mean validation accuracy, close to the best Full FT regime, while preserving substantially higher CIFAR-100 performance.

Oxford-IIIT Pets presents a more nuanced pattern. Full FT is contractive throughout the grid, with mean entropy shifts ranging from $-1.57\%$ to $-3.63\%$. LoRA is structurally milder, ranging from essentially neutral at $10^{-6}$ ($+0.01\%$) to moderately contractive at $5{\times}10^{-5}$ ($-0.75\%$). However, low-learning-rate LoRA substantially underfits the task: at $10^{-6}$ it reaches only $19.13\%$ mean validation accuracy despite retaining near-baseline zero-shot transfer. This underfitting regime explains why method-only averages can be misleading on Pets. Once the learning rate is large enough to optimize the task effectively, LoRA becomes competitive in-domain while retaining substantially more of the pretrained transfer behavior than Full FT. The best Pets LoRA regime in the matched grid ($5{\times}10^{-5}$) attains $91.91\%$ mean validation accuracy, slightly above the best Pets Full FT regime ($91.20\%$ at $10^{-5}$), while preserving substantially higher CIFAR-100 accuracy ($51.91\%$ versus $5.01\%$).

\begin{figure*}[t]
    \centering
    \includegraphics[width=\textwidth]{figures/fig_controlled_lr_panels.png}
    \caption{Controlled matched-learning-rate comparison across EuroSAT and Oxford-IIIT Pets. Each panel reports the mean and standard deviation across five seeds for best validation accuracy, CLS-to-patch entropy drift, and adapter-aware CIFAR-100 zero-shot accuracy. Learning rate changes both the direction and magnitude of attention drift, and LoRA preserves substantially more transfer than Full FT at matched optimization scales.}
    \label{fig:lr-panels}
\end{figure*}

\begin{table*}[t]
    \centering
    \caption{Aggregate means across the completed matched-learning-rate grid. Each row summarizes 20 runs (four learning rates, five seeds).}
    \label{tab:aggregate}
    \resizebox{\textwidth}{!}{%
    \begin{tabular}{llrrrr}
        \toprule
        Dataset & Method & Mean $\Delta$Entropy (\%) & Mean $\Delta$ERF (\%) & Mean Best Val Acc (\%) & Mean CIFAR-100 ZS (\%) \\
        \midrule
        EuroSAT & Full FT & -0.47 & -1.97 & 98.96 & 11.28 \\
        EuroSAT & LoRA r=8 & +1.18 & +1.55 & 96.59 & 45.13 \\
        Oxford-IIIT Pets & Full FT & -2.32 & -5.18 & 88.94 & 8.54 \\
        Oxford-IIIT Pets & LoRA r=8 & -0.33 & -1.44 & 70.76 & 58.01 \\
        \bottomrule
    \end{tabular}
    }%
\end{table*}

Table~\ref{tab:aggregate} provides a compact summary over the full grid. On EuroSAT, LoRA is both structurally broader and substantially more transfer-preserving than Full FT, albeit with a modest average in-domain accuracy penalty driven largely by the smallest learning rate. On Pets, the method averages should be interpreted with caution because the $10^{-6}$ LoRA regime is dominated by underfitting. Even so, the aggregate comparison preserves the same structural contrast: Full FT is markedly more contractive than LoRA.

\section{Transfer Retention and Layerwise Drift}
\subsection{Transfer Retention}
The completed controlled study also sharpens the relationship between structure and transfer. The pretrained CLIP baseline reaches $60.22\%$ zero-shot accuracy on CIFAR-100. Against that reference point, the mean transfer loss under Full FT is severe across both datasets: $11.28\%$ average CIFAR-100 accuracy over the EuroSAT grid and $8.54\%$ over the Pets grid. LoRA retains much more of the pretrained capability: $45.13\%$ on EuroSAT and $58.01\%$ on Pets.

Importantly, the run-level association between structural preservation and transfer preservation is stronger than the association between structural preservation and in-domain validation accuracy. The entropy-vs-CIFAR correlation is $0.61$ on EuroSAT and $0.90$ on Pets, whereas the entropy-vs-accuracy correlation is weak to moderate ($-0.14$ and $-0.43$, respectively). This pattern should be interpreted cautiously. It does not imply that broader attention support causes better zero-shot transfer, but it suggests that attention heatmap drift can serve as a descriptive proxy for how much of the pretrained representation is retained during adaptation.

\subsection{Layerwise Drift}
\begin{figure*}[t]
    \centering
    \includegraphics[width=\textwidth]{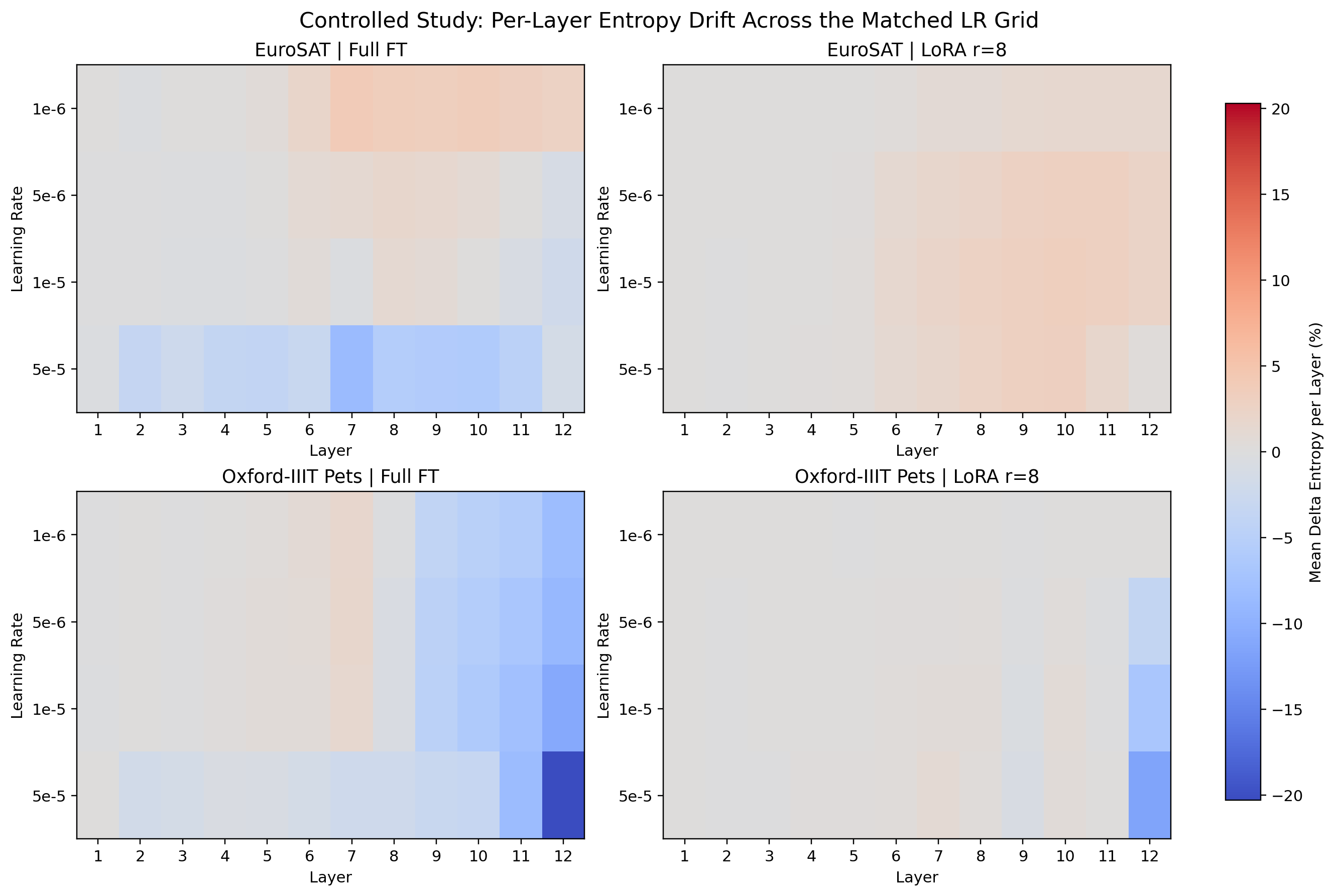}
    \caption{Mean per-layer entropy drift across the matched learning-rate grid. Rows correspond to learning rates and columns correspond to transformer layers. The strongest contraction is concentrated in the later layers, especially under high-learning-rate Full FT on Oxford-IIIT Pets, whereas LoRA remains structurally milder across the grid.}
    \label{fig:layer-heatmaps}
\end{figure*}

Figure~\ref{fig:layer-heatmaps} visualizes mean per-layer entropy drift over the matched grid. The layerwise view adds an important qualification to the run-level averages. Structural contraction is not uniformly distributed across depth. Under high-learning-rate Full FT, the largest negative shifts accumulate in the later layers, especially on Pets. At $5{\times}10^{-5}$, the mean layer-12 entropy shift is $-20.29\%$ for Full FT and $-11.53\%$ for LoRA. EuroSAT shows the same directional tendency under high-learning-rate Full FT, albeit less strongly.

LoRA does not eliminate drift, but it changes its profile. On EuroSAT, the LoRA heatmaps remain positive across all learning rates, with especially large broadening in the middle and later layers. On Pets, LoRA becomes mildly contractive as the learning rate increases, yet the contraction remains substantially smaller than under Full FT. This distinction supports a more precise interpretation of the adaptation effect: the dominant pattern is not uniform collapse across optimization settings, but pronounced late-layer contraction under aggressive Full FT regimes.

\section{Supplementary Analyses}
We include several supplementary analyses that broaden the empirical picture without changing the evidentiary hierarchy of the paper. The matched-learning-rate matrix remains the primary basis for the main claims, while rollout, patch-to-patch, CKA, backbone, regularization, and rank-ablation results test whether the same interpretation survives under additional views of the models.

\subsection{Structural Validation}
The first supplementary group triangulates the controlled entropy findings with additional structural measurements. Table~\ref{tab:auxiliary-validation} shows that on EuroSAT, LoRA r=8 produces slightly higher rollout entropy, larger rollout ERF@0.95, lower rollout Gini, and higher patch-to-patch entropy than Full FT. These auxiliary metrics are consistent with the controlled matched-learning-rate conclusion that LoRA is generally more structurally conservative than Full FT.

The same table also reports a low coefficient of variation across five balanced evaluation subsets, suggesting that the reported entropy measurements are not highly sensitive to the specific subset of images used for the structural analysis. This stability reduces the concern that the observed drift patterns are artifacts of evaluation-image selection.

Representational similarity provides an additional perspective. Table~\ref{tab:cka-entropy} shows that higher mean CKA to the pretrained encoder is associated with broader retained attention support ($r=0.585$ across 39 checkpoints). Runs that remain closer to the pretrained representation geometry also tend to preserve broader attention support. This result does not establish mechanism, but it is directionally consistent with the controlled entropy-versus-transfer correlations reported above \citep{kornblith2019cka}.

\begin{table*}[t]
\centering
\caption{Auxiliary validation analyses beyond CLS-to-patch entropy. Rollout and patch-to-patch metrics indicate broader attention support for LoRA than for Full FT on EuroSAT, while layerwise CKA \citep{kornblith2019cka} shows higher representational similarity to the pretrained encoder. The subset-sensitivity coefficient of variation (CV) is low for both methods, indicating that the reported metrics are stable across five stratified, class-balanced evaluation subsets drawn from the in-domain validation split.}
\label{tab:auxiliary-validation}
\small
\resizebox{\textwidth}{!}{%
\begin{tabular}{lrrrrrr}
\toprule
Model & Rollout Entropy (bits) & Rollout ERF@0.95 & Rollout Gini & Patch-to-Patch Entropy (bits) & Mean Layerwise CKA & Entropy CV \\
\midrule
Full FT EuroSAT & 5.593 & 0.944 & 0.090 & 4.721 & 0.694 & 0.0006 \\
LoRA r=8 EuroSAT & 5.605 & 0.952 & 0.064 & 4.794 & 0.815 & 0.0005 \\
\bottomrule
\end{tabular}
}%
\end{table*}

\begin{table}[t]
\centering
\caption{Correlation between representational fidelity (CKA to the pretrained model \citep{kornblith2019cka}) and structural change ($\Delta$entropy).}
\label{tab:cka-entropy}
\small
\begin{tabular}{lrrrrr}
\toprule
Analysis & $n$ & Pearson $r$ & Pearson $p$ & Spearman $\rho$ & Spearman $p$ \\
\midrule
Run-level mean CKA vs. mean $\Delta$Entropy & 39 & 0.585 & 0.0003 & 0.406 & 0.0106 \\
Strongest per-layer association (layer 4) & 39 & 0.892 & 0.0000 & 0.638 & 0.0000 \\
\bottomrule
\end{tabular}
\end{table}

\subsection{Backbone and Regularization}
The paper also includes two contextual checks. First, Table~\ref{tab:backbone-comparison} shows that backbone choice changes the magnitude and even the sign of structural drift on EuroSAT. On ViT-B/16, Full FT increases entropy where ViT-B/32 is nearly neutral or mildly contractive. This reinforces the broader caution from the controlled study: structural drift is not uniform across pretrained backbones.

Second, the regularization results show that explicit structural penalties can move Full FT toward broader attention support, but the evidence is not yet sufficient to support a central claim of reliable mitigation. Table~\ref{tab:regularization} indicates that APR and entropy-floor regularization can improve entropy relative to the unregularized Full FT baseline, and Table~\ref{tab:stats} shows that none of the tested regularizers survives family-wise correction in the current analysis \citep{holm1979multiple}. These results are therefore presented as follow-up experiments rather than as central evidence.

For Table~\ref{tab:stats}, the learning-rate correlations are evaluated with exact permutation tests because the matched grid contains only five learning-rate settings, and the regularizer comparisons are adjusted with Holm--Bonferroni across the regularizer family to control family-wise error \citep{holm1979multiple}.

At the run level, the same matched-grid pattern is supported by the multi-seed follow-up table: on EuroSAT, Full FT and LoRA r=8 differ significantly in mean entropy ($p=0.0127$), ERF ($p=0.0026$), Gini ($p=0.0368$), head diversity ($p=0.0430$), and best validation accuracy ($p=0.0699$ by Welch test); on Pets, the entropy and ERF differences remain significant ($p=0.0064$ and $p=0.0228$), as do Gini ($p=0.0019$) and best validation accuracy ($p=0.0115$). Accordingly, the principal EuroSAT/Pets method contrast is interpreted as statistically supported rather than exclusively descriptive \citep{holm1979multiple}.

\begin{table}[t]
\centering
\caption{Backbone comparison between CLIP ViT-B/32 and ViT-B/16 on EuroSAT. Metric cells show B/32 / B/16 values.}
\label{tab:backbone-comparison}
\small
\begin{tabular}{lrrrrr}
\toprule
Method & $\Delta$Entropy (\%) & $\Delta$ERF (\%) & $\Delta$Gini (\%) & Task Acc. (\%) & Entropy Gap \\
\midrule
Full FT on EuroSAT & -0.14 / +3.57 & -1.73 / +6.66 & -3.54 / -8.83 & 99.30 / 99.15 & +3.72 \\
LoRA r=8 on EuroSAT & +0.58 / -0.59 & +0.37 / +4.88 & -6.48 / -4.09 & 98.96 / 98.89 & -1.18 \\
\bottomrule
\end{tabular}
\end{table}

\begin{table}[t]
\centering
\caption{Regularization study on EuroSAT full fine-tuning. Among individually tested regularizers, entropy floor with $\lambda=0.1$ is the only setting with a paired per-layer entropy difference reaching $p<0.05$ in the current analysis.}
\label{tab:regularization}
\small
\begin{tabular}{lrrrr}
\toprule
Setting & Task Acc. (\%) & $\Delta$Entropy (\%) & $\Delta$ERF (\%) & $\Delta$Gini (\%) \\
\midrule
No regularizer & 99.30 & -0.14 & -1.73 & -3.54 \\
APR $\lambda=0.01$ & 99.26 & +0.86 & +1.50 & -8.83 \\
APR $\lambda=0.1$ & 99.19 & +0.56 & +1.03 & -5.36 \\
APR $\lambda=1.0$ & 98.96 & +0.09 & +0.16 & -0.75 \\
Entropy floor $\lambda=0.01$ & 99.11 & +0.58 & -0.09 & -6.41 \\
Entropy floor $\lambda=0.1$ & 99.07 & +0.79 & +0.30 & -8.22 \\
Weight decay 0.0 & 99.19 & -0.18 & -1.81 & -3.16 \\
Weight decay 0.1 & 99.15 & -0.25 & -1.90 & -2.38 \\
\bottomrule
\end{tabular}
\end{table}

\begin{table*}[t]
\centering
\caption{Run-level multi-seed comparisons on mean $\Delta$entropy. Values are reported as percentage change relative to the corresponding pretrained baseline, aggregated across completed seed runs.}
\label{tab:multiseed}
\small
\begin{tabular}{lrrrrr}
\toprule
Comparison & Seeds & Mean $\Delta$Entropy (\%) & Welch $t$ & Welch $p$ & Cohen's $d$ \\
\midrule
Full FT vs. LoRA r=8 (EuroSAT) & 3 vs 3 & -0.22 vs +0.87 & -7.157 & 0.0127 & -5.844 \\
Full FT vs. LoRA r=8 (Pets) & 3 vs 3 & -2.10 vs -0.84 & -7.969 & 0.0064 & -6.507 \\
Full FT vs. APR $\lambda=0.1$ (EuroSAT) & 3 vs 3 & -0.22 vs +0.55 & -18.441 & 0.0020 & -15.057 \\
Full FT vs. Entropy floor $\lambda=0.1$ (EuroSAT) & 3 vs 3 & -0.22 vs +0.70 & -14.439 & 0.0002 & -11.789 \\
\bottomrule
\end{tabular}
\end{table*}

\begin{table}[t]
\centering
\caption{Corrected inferential statistics.}
\label{tab:stats}
\small
\begin{tabular}{llrr}
\toprule
Comparison & Test & Statistic & $p$-value \\
\midrule
LR vs. $\Delta$Entropy & Exact permutation Pearson $r$ & -0.893 & 0.0333 \\ 
LR vs. $\Delta$Entropy & Exact permutation Spearman $\rho$ & -1.000 & 0.0167 \\ 
APR $\lambda=0.01$ vs. baseline & Paired per-layer t-test & 2.132 & 0.2194 \\ 
Entropy floor $\lambda=0.1$ vs. baseline & Paired per-layer t-test & 2.638 & 0.1154 \\ 
\bottomrule
\end{tabular}
\end{table}

\subsection{Single-Configuration Checks}
Some earlier single-configuration results remain informative when interpreted in light of the controlled matrix. In the original EuroSAT rank ablation, LoRA r=4, r=8, and r=16 all produced positive entropy shifts ($+0.98\%$, $+0.58\%$, and $+0.44\%$, respectively) and materially higher adapter-aware CIFAR-100 zero-shot accuracy than the corresponding original Full FT baseline (40.51\%, 47.28\%, and 39.35\% versus 9.43\%). These values do not override the controlled matched-learning-rate comparison, but they are consistent with the same overall interpretation: lower-rank adaptation tends to retain more of the pretrained structure and transfer behavior than unconstrained Full FT.

\section{Practical Implications}
The controlled study yields four practical implications for CLIP adaptation.
\begin{itemize}
    \item When transfer retention matters, adaptation methods should be compared at matched learning rates rather than through method-only averages.
    \item High-learning-rate Full FT is the regime most consistently associated with late-layer contraction and severe zero-shot degradation.
    \item Low-learning-rate LoRA can underfit the downstream task, so weak LoRA averages may reflect optimization regime rather than an intrinsic limitation of low-rank adaptation.
    \item Attention drift provides a lightweight descriptive diagnostic of representation preservation, especially when interpreted jointly with transfer metrics rather than as a stand-alone causal explanation.
\end{itemize}
These takeaways do not eliminate dataset or backbone dependence, but they provide a more actionable reading of the results than a generic ``attention collapse'' narrative.

\section{Limitations}
This study has four main limitations. First, the controlled matrix is restricted to CLIP ViT-B/32; although a ViT-B/16 auxiliary check is retained, the main controlled claims should not be generalized automatically across backbones. Second, the transfer analysis centers on CIFAR-100 as the only primary out-of-domain benchmark. Third, the paper compares Full FT only to LoRA r=8 in the controlled matrix; it does not provide exhaustive coverage of the broader parameter-efficient fine-tuning design space. Fourth, the supplementary analyses are heterogeneous in design and are therefore used to contextualize the controlled matrix rather than to replace it.

\section{Conclusion}
This paper presented a completed 80-run controlled comparison of Full FT and LoRA for CLIP ViT-B/32 under matched learning rates and used that matrix as the primary basis for interpreting attention drift and transfer retention. The results show that optimization scale is a first-order driver of structural change, that LoRA is generally more transfer-preserving than Full FT at matched learning rates, and that broader retained attention support is positively associated with better zero-shot retention. The supplementary analyses extend this interpretation without displacing it: rollout, patch-to-patch, and CKA summaries are directionally consistent with the controlled matrix, regularization results suggest partial but not yet definitive mitigation, and backbone validation shows that the magnitude of drift depends on architectural context. The overall conclusion is deliberately practical: matched-learning-rate evaluation is essential for credible Full-FT-versus-LoRA comparisons, and attention drift is most useful as a descriptive empirical lens on representation preservation in CLIP adaptation.

\clearpage
\appendix

\section{Reproducibility Details}

\subsection{Model, data, and optimization}
The released experiments use CLIP ViT-B/32 (\texttt{openai/clip-vit-base-patch32}) with eager attention extraction enabled in the vision encoder. The controlled matrix compares Full FT and LoRA $r=8$ on EuroSAT and Oxford-IIIT Pets over shared learning rates $10^{-6}$, $5{\times}10^{-6}$, $10^{-5}$, and $5{\times}10^{-5}$ and seeds 7, 11, 19, 42, and 123. The released training code uses AdamW with weight decay 0.01, a 5\% linear warmup followed by cosine decay, cross-entropy loss, and gradient clipping at norm 1.0. The main training loaders use batch size 64; fixed structural-evaluation subsets are processed at batch size 32. In the released scripts, EuroSAT runs are trained for 20 epochs and Oxford-IIIT Pets runs for 30 epochs, with best validation accuracy recorded across epochs rather than by early stopping.

Dataset loading in the released code uses Hugging Face datasets: EuroSAT (\texttt{tanganke/eurosat}), Oxford-IIIT Pets (\texttt{timm/oxford-iiit-pet}), CIFAR-100 (\texttt{uoft-cs/cifar100}), and Flowers102 (\texttt{nelorth/oxford-flowers}, supplementary only). Images are resized to $224\times224$, center-cropped, normalized with the CLIP mean and standard deviation, and augmented during training with random horizontal flips. Structural measurements are computed on fixed class-balanced subsets of 200 validation images produced by \texttt{create\_fixed\_eval\_subset}; the release also includes a subset-sensitivity check across five subset seeds.

For the main controlled comparison, LoRA is applied to the vision attention projections \texttt{q\_proj} and \texttt{v\_proj} with rank $r=8$, $\alpha=16$, and dropout 0.05; the classifier head and visual projection remain trainable. Supporting ablations vary the LoRA rank and, in one auxiliary check, expand the target modules to \texttt{q\_proj}, \texttt{k\_proj}, \texttt{v\_proj}, and \texttt{out\_proj}.

\subsection{Attention and transfer evaluation}
Attention maps are extracted with \texttt{output\_attentions=True} from the CLIP vision tower. For each layer, the analysis takes CLS-to-patch attention (token 0 attending to tokens 1:49), renormalizes over the 49 spatial patches, and computes entropy, ERF@0.95, Gini concentration, and head diversity. Supplementary rollout metrics average heads, mix each layer with a residual identity term, and compose the resulting matrices across depth. Patch-to-patch entropy excludes the CLS token and averages row entropy over patch-to-patch attention.

Adapter-aware CIFAR-100 zero-shot accuracy is the primary transfer metric. Text prompts use the template ``a photo of a \{class name\}'' and are encoded once with the CLIP text tower. Image features are extracted through the active adapted vision tower and visual projection, normalized, compared with normalized text features by dot product, and evaluated with top-1 accuracy on the CIFAR-100 test split. The supplementary release also includes Flowers102 zero-shot numbers. Importantly, the paper uses the adapter-aware path implemented in \texttt{followup\_analysis.py} for LoRA checkpoints rather than the older base-model-only loader.

\subsection{Released artifacts}
The manuscript package includes the following top-level artifacts: \path{main.tex}, \path{tables/}, \path{figures/}, and \path{paper_assets.json}. In the companion repository, run-level artifacts are stored as JSON histories under \path|outputs/metrics/*_history.json| together with checkpoints under \path|outputs/checkpoints/<experiment>/best_model.pth|. The public paper package summarizes these artifacts rather than duplicating every JSON file. The release stores run-level histories as JSON rather than CSV, and \path{scripts/build_assets.py} regenerates the manuscript tables from those metrics artifacts.

The released code supports both CUDA and Apple MPS backends. The companion English experiment report records an NVIDIA A40 (46~GB) environment for the earlier experiment bundle, while the checked-in runner also includes an MPS configuration for local reruns. Because hardware metadata are not logged uniformly for every artifact, we make no wall-clock or throughput claims in this paper.

\section{Additional Tables and Statistical Details}
For the small-$n$ learning-rate analysis, we report exact permutation correlations on the five-point learning-rate grid. Regularizer comparisons are adjusted with Holm--Bonferroni across the tested regularizer family, and the multi-seed follow-up contrasts are reported with Welch tests and effect sizes. Tables~\ref{tab:lr-sweep} and \ref{tab:zeroshot} provide exact supporting numbers for the learning-rate sweep and the adapter-aware zero-shot evaluation.

\begin{table}[t]
\centering
\caption{Learning-rate sweep for EuroSAT full fine-tuning. Entropy decreases monotonically as the learning rate increases across the tested range. Zero-shot columns report CIFAR-100 \citep{krizhevsky2009learning} and Flowers102 \citep{nilsback2008automated}.}
\label{tab:lr-sweep}
\small
\resizebox{\columnwidth}{!}{%
\begin{tabular}{lrrrrrr}
\toprule
LR & Task Acc. (\%) & $\Delta$Entropy (\%) & $\Delta$ERF (\%) & $\Delta$Gini (\%) & CIFAR-100 ZS (\%) & Flowers102 ZS (\%) \\
\midrule
1e-6 & 98.85 & +1.45 & +1.94 & -16.60 & --- & --- \\
5e-6 & 99.11 & +0.33 & -0.49 & -6.68 & --- & --- \\
1e-5 & 99.30 & -0.14 & -1.73 & -3.54 & 12.71 & 1.37 \\
5e-5 & 98.89 & -4.19 & -8.48 & +23.44 & 2.07 & 0.88 \\
1e-4 & 97.85 & -10.40 & -18.02 & +56.94 & 1.86 & 0.98 \\
\bottomrule
\end{tabular}
}%
\end{table}

\begin{table}[t]
\centering
\caption{Zero-shot results using an adapter-aware image-encoding path for LoRA checkpoints. The pretrained baseline is retained from the original CLIP evaluation path, while adapted checkpoints are re-evaluated with their own image branch active. Flowers102 is the standard flower-classification benchmark \citep{nilsback2008automated}.}
\label{tab:zeroshot}
\small
\begin{tabular}{lrr}
\toprule
Model & CIFAR-100 (\%) & Flowers102 (\%) \\
\midrule
Pretrained baseline & 60.22 & 0.20 \\
Full FT EuroSAT & 9.43 & 1.37 \\
LoRA r=4 EuroSAT & 40.51 & 1.27 \\
LoRA r=8 EuroSAT & 47.28 & 0.78 \\
LoRA r=16 EuroSAT & 39.35 & 0.98 \\
\bottomrule
\end{tabular}
\end{table}

\bibliographystyle{plainnat}
\bibliography{references}

\end{document}